\documentclass[sigconf]{acmart} 
\usepackage{multirow}
\usepackage{algorithm}
\usepackage{algpseudocode}

\AtBeginDocument{%
  \providecommand\BibTeX{{%
    \normalfont B\kern-0.5em{\scshape i\kern-0.25em b}\kern-0.8em\TeX}}}

\setcopyright{acmcopyright}
\copyrightyear{2024}
\acmYear{2024}

\acmConference[KDD'24]{ACM SIGKDD Conference on Knowledge Discovery and Data Mining}{August 25--29,
  2024}{Barcelona, Spain}
\begin{document}

\title{REFT: Resource-Efficient Federated Training Framework for Heterogeneous and Resource-Constrained Environments}

\author{Humaid Ahmed Desai}
\email{humaiddesai@vt.edu}
\affiliation{%
 \institution{Virginia Tech}
 \city{Blacksburg}
 \state{Virginia}
 \country{USA}}

\author{Amr Hilal}
\email{ahilal@vt.edu}
\affiliation{%
 \institution{Virginia Tech}
 \city{Blacksburg}
 \state{Virginia}
 \country{USA}}

\author{Hoda Eldardiry}
\email{hdardiry@vt.edu}
\affiliation{%
 \institution{Virginia Tech}
 \city{Blacksburg}
 \state{Virginia}
 \country{USA}}


\begin{abstract}
    Federated Learning (FL) plays a critical role in distributed systems. In these systems, data privacy and confidentiality hold paramount importance, particularly within edge-based data processing systems such as Internet-of-Things (IoT) devices deployed in smart homes. FL emerges as a privacy-enforcing sub-domain of machine learning that enables model training on client devices, eliminating the necessity to share private data with a central server. While existing research has predominantly addressed challenges pertaining to data heterogeneity, there remains a current gap in addressing issues such as varying device capabilities and efficient communication. These unaddressed issues raise a number of implications in resource-constrained environments. In particular, the practical implementation of FL-based IoT or edge systems is extremely inefficient. In this paper, we propose ``Resource-Efficient Federated Training Framework for
Heterogeneous and Resource-Constrained Environments (\emph{REFT})'', a novel approach specifically devised to address these challenges in resource-limited devices. Our proposed method uses Variable Pruning to optimize resource utilization by adapting pruning strategies to the computational capabilities of each client. Furthermore, our proposed REFT technique employs knowledge distillation to minimize the need for continuous bidirectional client-server communication. This achieves a significant reduction in communication bandwidth, thereby enhancing the overall resource efficiency. We conduct experiments for an image classification task, and the results demonstrate the effectiveness of our approach in resource-limited settings. Our technique not only preserves data privacy and performance standards but also accommodates heterogeneous model architectures, facilitating the participation of a broader array of diverse client devices in the training process, all while consuming minimal bandwidth.
\end{abstract}



\settopmatter{printfolios=true}
\maketitle

\section{Introduction}
Recent advancements in deep learning have yielded significant progress across diverse domains, including, but not limited to, image classification and natural language processing. Nevertheless, the training of complex Deep Neural Network (DNN) models necessitates the availability of massive amounts of data. Training models with substantial data volumes work well in centralized scenarios where the model has access to all of the data. However, in most cases, particularly within distributed systems involving Internet-of-Things (IoT) and edge devices, data resources are inherently decentralized. This decentralized data distribution presents challenges for collaborative training, primarily stemming from technical intricacies, privacy concerns, and the intricacies of data ownership. In response to these challenges, Federated Learning (FL) has emerged as a promising solution, enabling distributed model training through decentralized data while maintaining data privacy.

FL techniques facilitate collaborative training by iteratively sharing model parameters, or gradients, during the training process. This communication takes place between client devices connected to the Internet, encompassing IoT and edge devices, and a central server located remotely in the cloud, thus forming an FL-based system. Typically, this exchange is carried out via widely adopted protocols such as TCP or UDP \cite{vineeth_2021}, or alternative application layer protocols \cite{FedComm}. After each round of training on local data, a client transmits its model parameters to the central server, which then aggregates these parameters from all clients using traditional data aggregation methods. However, this iterative process necessitates a significant number of back-and-forth client-server communications, leading to increased bandwidth consumption and reduced communication efficiency. This challenge poses significant hurdles in the establishment of efficient FL-based systems, particularly in resource-constrained environments, such as smart home setups involving IoT and edge devices with limited hardware resources and simplified web infrastructure. Effectively harnessing FL in such applications while giving equal importance to data privacy is challenging. To alleviate communication bottlenecks in such FL scenarios, certain research initiatives \cite{FedComm, vineeth_2021} have been dedicated to streamlining communication time and minimizing packet loss by optimizing application layer protocols. Specifically, they explore Message Queue Telemetry Transport (MQTT), Advanced Message Queuing Protocol (AMQP), and ZeroMQ Message Transport Protocol (ZMTP) to enhance the efficiency of data exchange.

Despite these advancements, challenges persist in FL-based systems, including the substantial number of bidirectional communication rounds, the size of each model update, and concerns related to data privacy. This leads us to the following questions: How can we execute FL with the fewest communication rounds, minimizing bandwidth consumption in each exchange and optimizing overall network traffic, all while preserving FL's fundamental principle of data privacy? Furthermore, client devices in FL, such as IoT and edge devices, are often significantly more constrained resources compared to data center servers, with limitations in processing power, memory, and storage. This prompts the subsequent question: How can we efficiently execute FL as IoT and edge devices become increasingly prevalent while optimizing resource usage on each client and accommodating a wide array of diverse client devices, particularly when working with complex DNN models?

To address these challenges, we propose ``Resource-Efficient Federated Training (REFT)'', a framework that combines variable pruning and knowledge distillation techniques. Variable pruning reduces model parameters based on the client's computational capacity, while knowledge distillation enhances communication efficiency and data privacy. Our method is inspired by FedKD \cite{FedKD}, which employs a public dataset to mitigate privacy concerns. Unlike existing methods, we perform asynchronous updates and leverage public data, reducing communication and privacy risks. Our experiments demonstrate significant reductions in parameters, FLOPs, and bandwidth consumption while maintaining accuracy levels comparable to existing FL techniques. We summarize our key contributions as follows:

\begin{itemize}
    \item We introduce \textbf{\textit{variable pruning}}, a framework for applying model pruning techniques that adjust the pruning level for individual clients based on their available computational resources. By performing one-shot structured pruning on the initial model weights at the server, we customize the pruning level for each client. This approach utilizes client resources efficiently and reduces both computational and communication overhead (or keeps them at an acceptable level), making the model more suitable for training on resource-constrained devices.

    \item We employ a one-way, one-shot client-to-server knowledge distillation approach using unlabeled, non-sensitive public data. This technique further enhances communication efficiency in federated learning by optimizing the transfer of knowledge from clients to the server. We also accommodate clients with heterogeneous model architectures, which are obtained after structured pruning, enabling their active participation in the training process.

\end{itemize}
Overall, our proposed approach combines variable model pruning and one-shot knowledge distillation to improve the efficiency and effectiveness of federated learning, making it more feasible for resource-constrained devices, resulting in better resource utilization on the client side, and accommodating diverse client architectures.

\section{Related Works} \label{related works}
\subsection{Efficient Federated Learning}
The foundational concept of Federated Learning by McMahan et al. \cite{McMahan} addresses decentralized training while safeguarding data privacy. Their approach, Federated Averaging (FedAvg), calculates local gradients on client data in each round, followed by parameter averaging via a server until convergence. Various FedAvg derivatives have emerged to address aspects like non-IID data \cite{FairAllocFL, AgnosticFL} and to introduce novel aggregation techniques \cite{FedVC, MatchedAvg}. However, high training costs persist as a challenge.

Recent efforts focus on curbing communication costs \cite{RobCommEFL, WirelessEnergyEFL, PruneFL, FedPAQ_PMLR, FL-PQSU}. While some \cite{sketched-updateFL} target client-to-server communication expenses, they overlook downlink (server-to-client) communication costs. Approaches like \cite{FL-PQSU} emphasize bit reduction during training through quantization and pruning, while also minimizing server communication for model updates. However, communication frequency with the server remains unchanged, limiting efficiency gains. Our REFT excels in communication and computation efficiency via pruning alone, surpassing \cite{FL-PQSU}, which combines quantization, pruning, and selective updates. The incorporation of knowledge distillation further enhances the efficiency of our approach.

\subsection{Model Compression} \label{Model Compression}
Neural network pruning optimizes large networks by removing redundant or irrelevant connections. Early attempts used second-order Taylor expansion \cite{LeChun1990, Hassibi1993}, but the impractical computation of the Hessian matrix led to alternative strategies. Han et al.'s work \cite{DeepCompression, Han2015} popularized magnitude-based pruning, where small-magnitude parameters are pruned. \cite{PruneFL, FL-PQSU} have leveraged pruning in federated learning to reduce communication and computation overhead. \cite{PruneFL} adopts a two-stage approach, initially unstructured pruning on a selected client and then "adaptively" pruning the model during learning, i.e., reconfiguring the model by removing and adding back parameters. However, unstructured pruning leads to irregularly sparse weight matrices and relies on weights to be stored in a compressed format. Consequently, such matrices are less compatible with data-parallel architectures in GPUs and multicore CPUs, necessitating specialized hardware and software support \cite{NonStruct-pruning}.


In contrast, \cite{FL-PQSU} employs structured pruning based on the L1 norm, combined with quantization and selective updates. Unlike unstructured pruning, structured pruning produces hardware-friendly weight matrices. However, for effective pruning benefits, it's crucial to reshape the weight matrices, reducing inference latency and model size. While \cite{FL-PQSU, PruneFL} rely on simulations to assess pruned model performance, creating weight masks to estimate potential reductions in model or parameter size, these approaches may not guarantee actual reductions. Our approach compresses models by reshaping input and output tensors using generated masks, aiming to reduce model complexity and enhance inference latency. Notably, \cite{FL-PQSU, PruneFL} concentrate solely on uniformly distributed data scenarios (for CIFAR10) which might not represent real-world FL settings characterized by non-IID data distribution. Such situations pose optimization challenges \cite{FLChallenges}. In contrast, our method is tailored to remain robust against data and model heterogeneity, ensuring convergence with superior or comparable performance.

\begin{figure*}[t]
    \centering
    \includegraphics[width=1\textwidth]{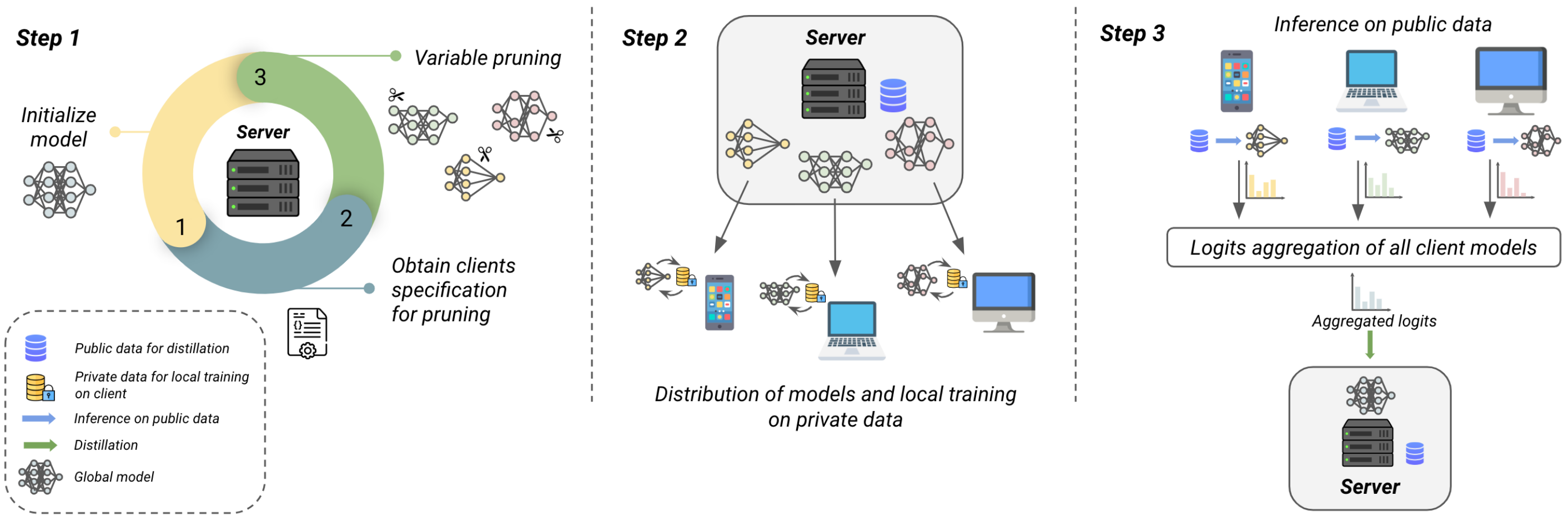}
    \caption{Overview of \emph{REFT}}
    \label{fig:REFTFramework}
    \Description{Overview of the three-step pipeline of REFT}
\end{figure*}

\subsection{Knowledge Distillation}
Knowledge distillation, initially introduced by Hinton et al. \cite{hinton2015distilling}, has witnessed significant progress in the realm of model ensemble, with a particular emphasis on the student-teacher learning paradigm, where the student model seeks to approximate the output logits of the teacher model \cite{romero2015fitnets, AttentiontoAttention, Tung_2019_ICCV}. Existing works either aim to average the logits from an ensemble of teacher models or extract knowledge at the feature level. The majority of these approaches utilize existing training data for the distillation process. Some works \cite{Zero-ShotKD, Zero-shotKTviaBM} have explored distillation through pseudo-data generation from the weights of the teacher model or through a generator adversarially trained with the student model, particularly when real data are unavailable for training. FedDF \cite{FedDF}, on the other hand, utilizes an unlabeled dataset for ensemble distillation, which is generated from a pre-trained GAN \cite{GAN}.

Guha et al. \cite{Guha2019} proposed a one-shot federated learning approach, wherein the server learns a global model of devices in the federated network in a single communication round. However, unlike their approach, which employs unlabeled public data collected from the same domain, we adopt the approach used by Gong et al. \cite{FedKD, gong2022FedAD, gongICCV}, which aggregates local predictions on unlabeled public data from different domains for enhanced privacy guarantee. Unlike Gong et al., whose primary focus is on preserving privacy, our approach is oriented towards increasing communication efficiency and resource utilization while maintaining privacy.

\section{Preliminaries} \label{preliminaries}
A typical Federated Learning system consists of a central server \emph{S} and a group of participating clients \emph{C}. Each client in the network possesses its own labeled private dataset $D_C$, which it uses to train a local model $M_C$. The server $S$ then combines these local updates using an aggregation method to obtain an updated global model, $M_G$. This iterative process continues until a stopping condition is met. The widely used FL algorithm, FedAvg \cite{McMahan}, initially defines a training task, including setting hyperparameters, and selects a fraction of clients for training. The initial global model, $M^0_G$ is then broadcast. In each round $R$, clients $C$ perform a local computation on their respective data and update their model parameters $W^R_C$. The primary objective is to minimize the global loss, which is a weighted average of the individual client's loss function.

\begin{equation} \label{eq:1}
    L(w) = \sum^C_{c=1} \frac{k_c}{k} \ell_c(w),
\end{equation}
\[\text{where } \ell_c(w) = \frac{1}{k_c} \sum_{i \in D_c} f_i(w)\]

Here, $k_c$ represents the number of data samples in the client's dataset $D_c$, $k$ denotes the total number of samples, and $\ell_c$ is the loss function of client $c$.

Ensuring constant communication with the server is essential to achieving the objective in Equation \ref{eq:1}. However, in the context of IoT and edge devices, these devices are often resource-constrained, which significantly limits communication and computation resources. Training DNNs on such devices becomes time-consuming, and multiple rounds of communication with the server result in substantial bandwidth consumption, making this approach inefficient for FL-based systems.

\section{Proposed Framework: \emph{REFT}}
Our aim is to lower communication and computation overheads within the FL process, with a specific focus on resource-constrained devices. These devices, which often fall within the IoT or edge-device paradigm, play a pivotal role in generating valuable data for the FL process.
However, existing pruning methods such as \cite{FL-PQSU, PruneFL} adopt a one-size-fits-all approach to pruning models without considering individual clients' hardware capabilities. This results in inefficient resource utilization among participating clients. To tackle these issues, we introduce the \emph{REFT} framework. It features a three-stage pipeline designed to decrease the training cost of complex DNNs, enhance client training's resource utilization, and optimize communication efficiency. This is achieved by minimizing redundant communications between clients and the server while maintaining data privacy.

In the first stage, we estimate the computational capacity of each client device by obtaining their FLOPS (Floating Point Operations Per Second) values and subsequently prune the model to a level suitable for training on that client. 
This approach presents two advantages: it reduces the model's complexity, thereby decreasing its computational demands, and it minimizes bandwidth usage during the initial model transfer.

The second stage encompasses the distribution of the global model, followed by individual client model training on their respective private datasets. To maintain privacy, the server is restricted to accessing only public data. Notably, the server maintains its own public dataset, which we presume is universally available to all clients as an independent dataset. Consequently, for the purpose of bandwidth calculation, we omit the inclusion of this server-side public dataset, focusing solely bandwidth cost of server-client communication during FL training. This exclusion aligns with the established practices in existing works, as exemplified in \cite{FedKD,FedDF}, wherein public datasets have been utilized. However, these studies have typically disregarded the associated communication costs of public datasets in their analyses. 

In the final stage, we employ knowledge distillation. This method curbs the need for frequent and large model updates between clients and the server, significantly cutting communication costs. Additionally, this distillation-based approach enhances the framework's versatility, allowing clients to possess distinct architectures aligned with their local data distribution and computational capabilities.

The details of the 3-stage REFT training process are depicted in Figure \ref{fig:REFTFramework} and are further described below:
\begin{itemize}
    \item \emph{Variable model pruning}:\
    In the initial round (round 0), the server $S$ initializes the global model parameters $W_G$ and prunes them according to client specifications. The pruned global model, $M^{0}_{G_P}$, is then sent to clients. Pruning is done just once using a one-shot approach, with pruning percentage $P$ ($100\% > P \geq 0\%$) varying based on client computational power. This method supports heterogeneous model architectures, allowing different clients to have distinct versions of $M^{0}_{G_P}$.
    
    \item {\emph{Model distribution and local training}}: \
    After pruning, the pruned model $M^{0}_{G_P}$ is broadcast along with the public dataset $D_P$. Each client $C$ then trains the model on its private labeled dataset $D_c = \{(x^i_c , y^i_c)\}$ where $i = 1,2,3,..., |D_c|$, and initializes the received global model $M^{0}_{G_P}$ with parameters $W_c$. It is important to note that the model architecture for each client may differ due to variable pruning or each client having its own custom model architecture.
    
    \item {\emph{Knowledge distillation}}: \
    To ensure privacy, the private datasets of clients are isolated. The public dataset $D_P$ on the server is employed for client-to-server knowledge distillation. An ensemble of local models $M_C$ and the global model $M_{G_P}$ forms a teacher-student arrangement. This maintains privacy while transferring knowledge from clients to the server.
\end{itemize}

\subsection{Variable Pruning} \label{Variable Pruning}
Training complex DNNs on resource-limited devices is often impractical due to their intricacy. Pruning, as explained in Section \ref{Model Compression}, provides a means to accelerate training and reduce computational demands. However, the inherent diversity of devices in federated learning adds complexity. Clients possess varying computational power, introducing challenges for traditional static pruning techniques. These methods, aimed at minimizing communication costs, employ uniform pruning strategies, irrespective of the diverse capacities of individual clients. Consequently, this approach underutilizes the potential of more capable clients by tailoring pruning to the least powerful client, failing to harness their potential for an efficient model
training.

Our approach addresses these limitations via variable-structured model pruning, tailored to individual client computational abilities. Unlike static pruning, which underestimates more potent clients, our strategy optimizes both computation and communication overheads. By assessing a client's computing capacity and estimating the necessary FLOPs for effective model training, we customize the pruning process. This results in models that are finely tuned for each client, enhancing the overall efficiency of the FL process. Given the significance of FLOPs in hardware assessment for DNN training \cite{FLOPsRTang2018, PowerComsumptionRTang2017}, we prioritize FLOPS as the primary metric for computation assessment and pruning degree determination. FLOPS provides a reasonable and hardware-independent measure for assessing the feasibility of neural network training. As memory capacity and other client hardware parameters can also offer valuable insights, we complement this assessment with an analysis of memory requirements (RAM) and GPU utilization for the DNNs discussed in the following sections.
We adopt L1 norm-based pruning, a straightforward method to gauge weight importance in DNNs \cite{li2017pruning}, and apply it to both convolutional and fully connected layers. This approach is justified as convolutional layers typically contribute significantly to computational overhead, while fully connected layers primarily impact the model size \cite{li2017pruning, PlayAndPrune}.

To effectively reduce the model's size and improve both training and inference speed, we employ the NNI toolkit. This toolkit allows us to perform pruning not only based on performance metrics like the L1 norm associated with each output channel but also to take into account the broader network architecture and its topology.  Specifically, we harness the toolkit's dependency-aware pruning technique \cite{nniOverviewModel} to identify and prune output channels shared by layers that exhibit channel dependencies. This ensures that the pruning process is carried out in a manner that preserves these critical inter-layer dependencies. The pruner (L1 norm algorithm) initiates the model pruning process, generating a weight mask. This mask is then subsequently utilized by the ModelSpeedup module to reconfigure the weight tensors, ensuring a meaningful reduction in both model size and inference speed. The overall variable pruning procedure is described in Algorithm \ref{Variable Pruning Algorithm}.


\subsection{Knowledge Distillation} \label{Knowledge Distillation}
In the knowledge distillation phase, we initiate local model training $M_c$ with the private labeled dataset $D_c$ at each client. Post-local training, the server dispatches an unlabeled public dataset $D_p = {(x^i_p)}$, with $i = 1,2,3,...,|D_p|$, to every client for knowledge distillation. Referring to \cite{FedKD}, the private dataset $D_c = {(x^i_c, y^i_c)}$, $i = 1,2,3,..., |D_c|$ (with $c \in C$), entails existing classes $T_c \subset {1,2,...,T}$ ($T$ as total classes across clients). The local model's output on the public data sample $x^i_p$ for class $t \in T_c$ is $z^i_{tc} = f(x^i_p, w_c, t)$, with $w$ representing model parameters. In high-data heterogeneity settings, traditional aggregation methods averaging all teachers' logits lack suitability, as the client’s dataset may not share identical target classes. To address this, we introduce the importance weight $I$ for each client, reflecting local private data distribution. The weight is computed as the ratio of samples in local client $c$ belonging to class $t$ to total samples across clients:
\begin{equation} \label{eq:w}
    I^t_c = \frac{N^t_c}{\sum_{c \in C} N^t_c},
\end{equation}
where $N^t_c$ is the sample count in local client $c$ for class $t$. It is important to note that this distillation holds true as long as the target class $t$ of public data sample $x^i_p$ matches the target class $T_c$ of the client's private dataset.

Kullback-Leibler divergence is utilized for teachers' soft label aggregation. Loss function $L$ is the cross-entropy sum between teacher and student model predicted probabilities. Here, $p_t$ and $q_t$ are probabilities of a sample belonging to class $t$ by teacher and student models:
\begin{equation} \label{loss-teacher-student}
    L = \sum_{t} p_t \log \frac{p_t}{q_t}
\end{equation}

The central model's output logits $\tilde{z}_t = f(x_p, w_s, t)$ form student knowledge, and aggregated logits $\hat{z}_t$ are teacher knowledge. Probabilities $p_t$ and $q_t$ are computed using softmax on logits with a temperature parameter $\tau$. As indicated in \cite{hinton2015distilling}, minimizing the loss with high-temperature parameter $\tau$ equates to minimizing the L2 norm between teacher and student network logits, simplifying loss to $L = ||\tilde{z}-\hat{z}||$.

REFT employs a one-shot offline distillation, predicting with each public data sample once, iteratively training the central model. This boosts privacy, reduces queries to local models, and limits local knowledge exposure. Moreover, synchronous updates and repetitive communication are eliminated, enhancing communication efficiency and flexibility.

\section{Variable Pruning vs. Static Pruning for Resource Utilization}
In the context of federated learning with $n$ clients ($c_1, c_2, ..., c_n$), each client ($c_i$) possesses unique hardware capability $h_{c_i}$ and computation capacity $F_{c_i}$ in FLOP. Our analysis centers on variable pruning versus static pruning's efficiency in harnessing resources. The pruning level $P_{c_i}$ for a client depends on its hardware, with $NP_{c_i} = N \times (1 - P_{c_i})$ total pruned model parameters, where $N$ is the total number of parameters in the unpruned model. We assume uniform pruning across layers and connections and a uniform distribution of FLOPs across pruned and unpruned parameters.

By reducing the number of parameters, we can reduce the training overhead and communication load. We define the FLOP reduction factor for client $c_i$ as $FP_{c_i} = F \times (1 - P_{c_i})$. The accuracy for training the pruned model is denoted as $AP_{c_i}$. It is worth noting that typically $A_{c_i} \geq AP_{c_i}$, where $A_{c_i}$ is unpruned model accuracy.

Static pruning ($P_{\text{static}}$) uses $P_{\text{min}}$ based on least capable hardware ($h_{\text{min}} = \min(h_1, h_2, ..., h_n)$). All clients, including those with higher hardware capabilities, are then pruned at the level $P_{\text{static}} = P_{\text{min}}$, resulting in $NP_{\text{static}} = N \times (1 - P_{\text{static}})$ total number of parameters. Thus, higher-capability clients with $F_{c_i} > F_{\text{static}}$ might be underutilized, leading to reduced training performance. This underutilization results in the following inequality: $A_{c_i} \geq AP_{c_i} \geq AP_{\text{static}}$, where $1 \geq P_{\text{static}} \geq P_{c_i} \geq 0$
We can define the utilization factor for client $c_i$ as $U_{c_i} = \frac{F_{\text{static}}}{F_{c_i}}$, indicating the ratio of the least performing client's hardware capacity to the client $c_i$'s hardware capacity. Our variable pruning strategy overcomes this underutilization by identifying a link between pruning level and FLOP reduction, which can be expressed as:
\begin{equation} \label{variable-pruning-equation}
    P_{c_i} = 1 - \frac{F_{c_i}}{F_{\lambda}}
\end{equation}
Here, $F_{c_i}$ signifies client $c_i$'s computational capability (FLOPS), and $F_{\lambda}$ is a trade-off coefficient between communication efficiency and accuracy. The choice of $F_{\lambda}$ allows the administrator (or orchestrator of FL) to prioritize either more efficient communication or better accuracy/performance. Note that if $F_{c_i} \geq F_{\lambda}$, no pruning will be performed for client $c_i$.

For instance, consider 5 clients with FLOP capacities of 10, 20, 40, 60, and 100 GFLOPS (GigaFLOPS). By setting $F_{\lambda}$ to 100 GFLOPS, pruning percentages are: $P_{c_1} = 90\%$, $P_{c_2} = 80\%$, $P_{c_3} = 60\%$, $P_{c_4} = 40\%$, and $P_{c_5} = 0\%$ (no pruning). This choice reflects the administrator's preference for communication efficiency over performance. Alternatively, opting for $F_{\lambda} = 50$ GFLOPS, clients $c_1$, $c_2$, and $c_3$ are pruned to $80\%$, $60\%$, and $20\%$ respectively, while clients $c_4$ and $c_5$ remain unpruned. This choice reflects the administrator's emphasis on accuracy or performance over communication costs.

\begin{algorithm}
\caption{Variable Pruning}\label{Variable Pruning Algorithm}
\begin{algorithmic}
    \Require{$n$ clients, each client $c_i \in C$ has computation capacity of $F_{c_i}$ FLOPS where $i = 1,..,n$ }, trade-off coefficient $F_\lambda$, client model $M_{c_i}$
    \For{each client $c_i$ in $C$}
        \State{$F_{c_i} \gets$ Request client's estimated FLOPS}
        \State{Calculate pruning ratio:} \State{$P_{c_i} \gets 1 - \frac{F_{c_i}}{F_{\lambda}}$} \Comment{Eq. 4}
        \State{Apply L1 Norm pruning and create weight mask:}
        \State{$m_{c_i} \gets L1NormPruner(M_{c_i}, P_{c_i})$}
        \State{$ModelSpeedup(M_{c_i}, m_{c_i})$} \Comment{Model reconfiguration}
    \EndFor
\end{algorithmic}
\end{algorithm}

\section{Experiments}
Our experiments aimed to assess our proposed approach's performance in image classification using the CIFAR10 and CIFAR100 datasets \cite{krizhevsky2009learning}. Private datasets for local training were created using a Dirichlet distribution, generating heterogeneous data splits \cite{non-iid-data, yurochkin2019bayesian}. The $\alpha$ parameter controlled dataset non-IID-ness, with higher values promoting similar data distributions across clients.

We employed simulations to emulate the training of diverse clients on HPC clusters. To accomplish this, we gathered estimated FLOPS values for various potential client devices and integrated these values into our simulation framework. For instance, devices in the category of Raspberry Pi models 3 and 4, and similar counterparts, were categorized as weak clients due to their FLOPS capabilities falling within the range of 8 to 40 GFLOPS. In a similar context, wearable devices, including high-end smartwatches and mobile phones, were classified as moderate to good clients, exhibiting FLOPS capacities below 150 GFLOPS. Conversely, devices equipped with high-performance computing units, such as laptops and desktops featuring dedicated GPUs, were designated as strong clients. It is noteworthy that our hardware configuration encompassed two 12 GB NVIDIA GRID P40-12Q GPUs and an Intel (R) Xeon (R) CPU E5-2640 v4 clocked at 2.40GHz.

\begin{figure}[t]
    \includegraphics[width=1\linewidth]{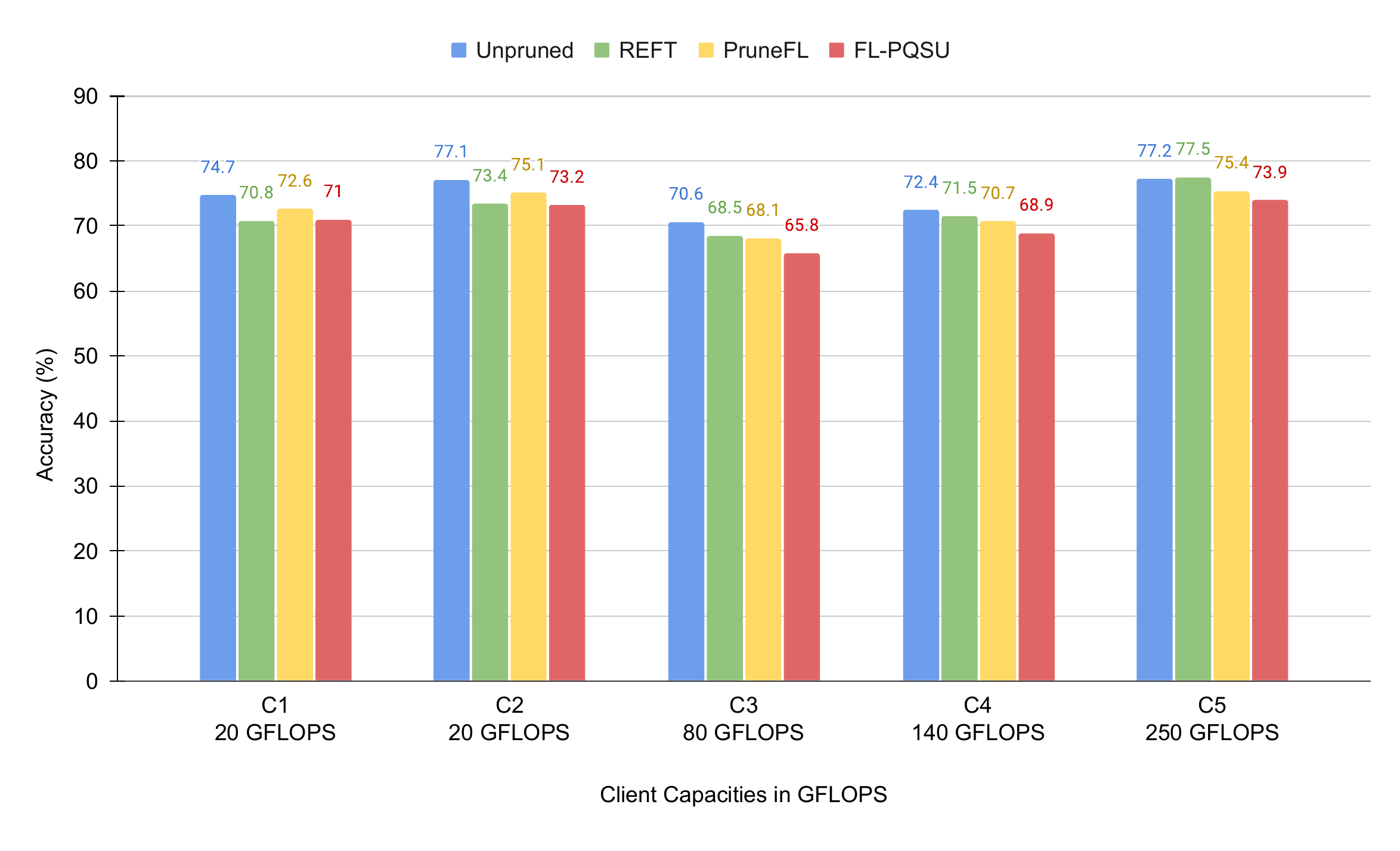}\centering
    \caption{Comparing client accuracies: static pruning (FL-PQSU and PruneFL) vs. REFT on the VGG-16 model.}\label{fig:comparision-accuracy}
\end{figure}

\subsection{Datasets and models}
For consistency and meaningful comparison, our experimental setup adhered to FedKD \cite{FedKD}. CIFAR10 was the private dataset, and CIFAR100 served as the public distillation dataset. While our experiments were exclusively conducted on these datasets, it is worth noting that our approach exhibits versatility and can potentially be adapted to accommodate different datasets, including but not limited to large-scale datasets like ImageNet, provided that the prerequisites outlined in Section \ref{Knowledge Distillation} regarding the private and public dataset conditions are met. We evaluated the test accuracy of our framework and compared it with the baselines. For our experiments, we employed ResNet-8 and VGG-16 model architectures, as described in the respective prior works \cite{FedKD, FL-PQSU}. To demonstrate the robustness of our approach, we performed experiments on non-IID data and created disjoint training sets for each client with the value of $\alpha$ set to 1.0.

\subsection{Baselines}
Our focus lies in enhancing resource utilization while maintaining performance and minimal communication. Thus, we compared against FedAvg \cite{McMahan}, PruneFL \cite{PruneFL}, FedKD \cite{FedKD}, and FL-PQSU \cite{FL-PQSU}. Personalized FL methods (Per-FedAvg \cite{PerFedAvg}, q-FedAvg \cite{q-fedAvg}) were excluded, as they prioritize adapting to individual client's data distribution.

\subsection{Metrics}
Our experiments focus on evaluating model performance in terms of test accuracy and comparing it with baseline approaches. Our main goal is to showcase resource-efficient training, achieving significant reductions in model size, parameter count, and FLOPs while maintaining minimal accuracy loss. We also delve into communication bandwidth analysis, a critical factor in federated learning that affects time and cost. For non-distillation training, communication bandwidth is computed based on parameters such as parameter size ($W$), the number of participating clients ($C$), communication rounds ($R$), and the number of bits ($B$) used for representing parameters and logits. This calculation is expressed as:
\begin{equation}\label{BandwidthCost}
\text{Bandwidth} = C \times R \times W \times B
\end{equation}

We analyze both downstream (server-to-client) and upstream (client-to-server) communication per client per round. For response-based knowledge distillation, where $W$ isn't transferred, we adapt the equation as $Bandwidth = L \times S \times B$, with $L$ as logits and $S$ as distillation steps.

\subsection{Implementation Details}
For ResNet-8, we used SGD optimizer with momentum of 0.9, weight decay of $3\times10^{-4}$, and Cosine Annealing scheduler (learning rate, $lr$, decreased from 0.0025 to 0.001) over 500 epochs with a batch size of 16. VGG-16 employed a $lr$ of 0.1 without weight decay and a batch size of 128. Optimizer, momentum, and epochs matched ResNet-8. Distillation employed a constant $lr$ of $10^{-3}$ and batch size of 512 with Adam optimizer.

For a fair comparison, we aligned with FedKD's hyperparameters for client model training. It utilized the SGD optimizer (momentum 0.9, weight decay $3\times10^{-4}$) with the Cosine Annealing scheduler over 500 epochs and batch size 16. Distillation used a constant $lr$ of $10^{-3}$ and batch size of 512 with Adam Optimizer.

\section{Results}

\begin{figure}
    \includegraphics[width=1\linewidth]{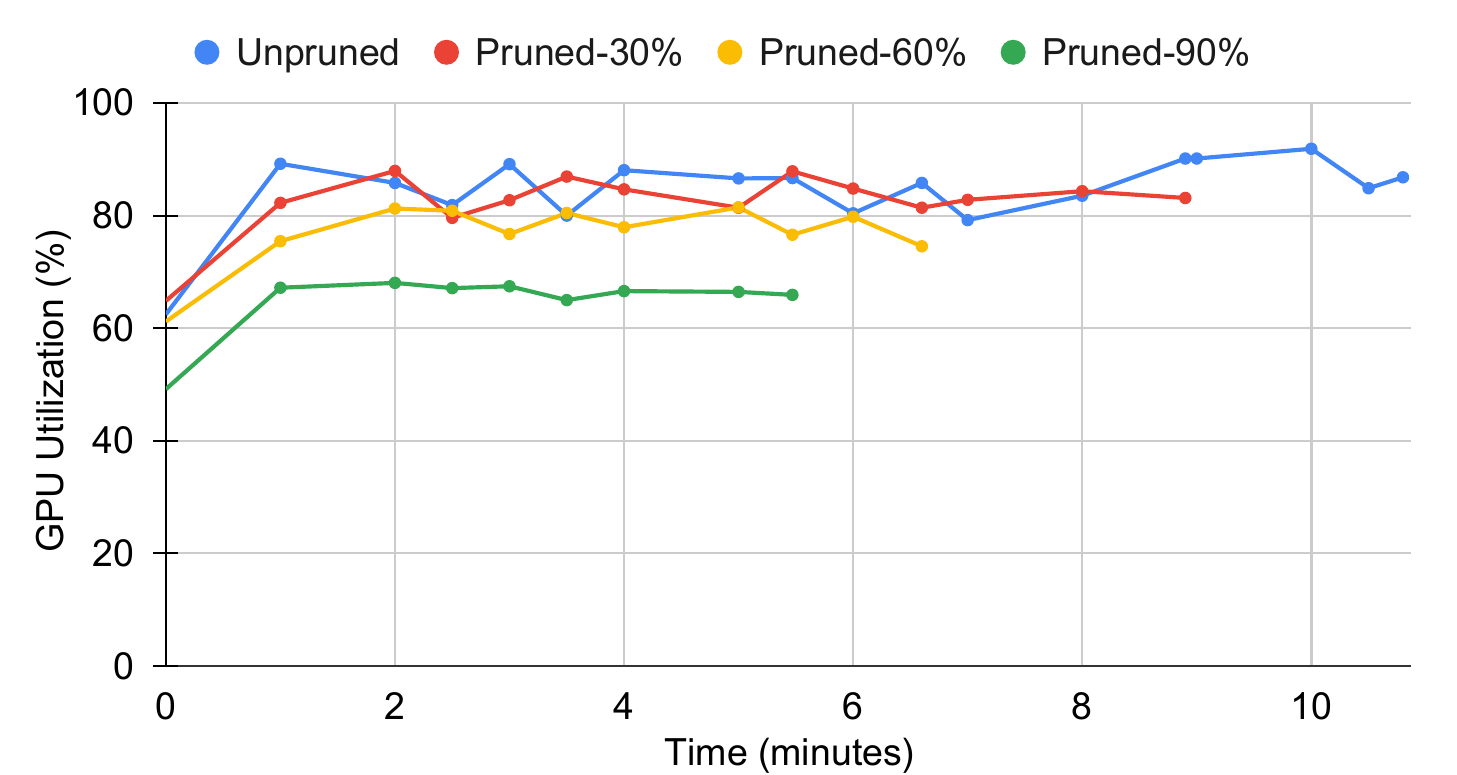}\centering
    \caption{Effect of pruning on GPU utilization and training time of VGG-16.}\label{fig:gpu-utilization}
\end{figure}

\subsection{Resource Utilization}
To demonstrate the effectiveness of our variable pruning strategy in optimizing hardware resource utilization, we compared accuracies per client across different pruning strategies (Figure \ref{fig:comparision-accuracy}). This experiment involved five clients with varying hardware capabilities. Clients $c_1$ and $c_2$ had limited computational capacity, necessitating a high pruning level (90\%) to accommodate the model. Pruning levels were determined using Eq. (\ref{variable-pruning-equation}), with $F_{\lambda}$ set to 200 GFLOPS ($10^9 \times$ FLOPS). Clients $c_3$ and $c_4$ possessed better hardware capabilities and required 60\% and 30\% pruning, respectively, for model training. Client $c_5$ had ample hardware resources and required no pruning.

\begin{figure}
    \includegraphics[width=1\linewidth]{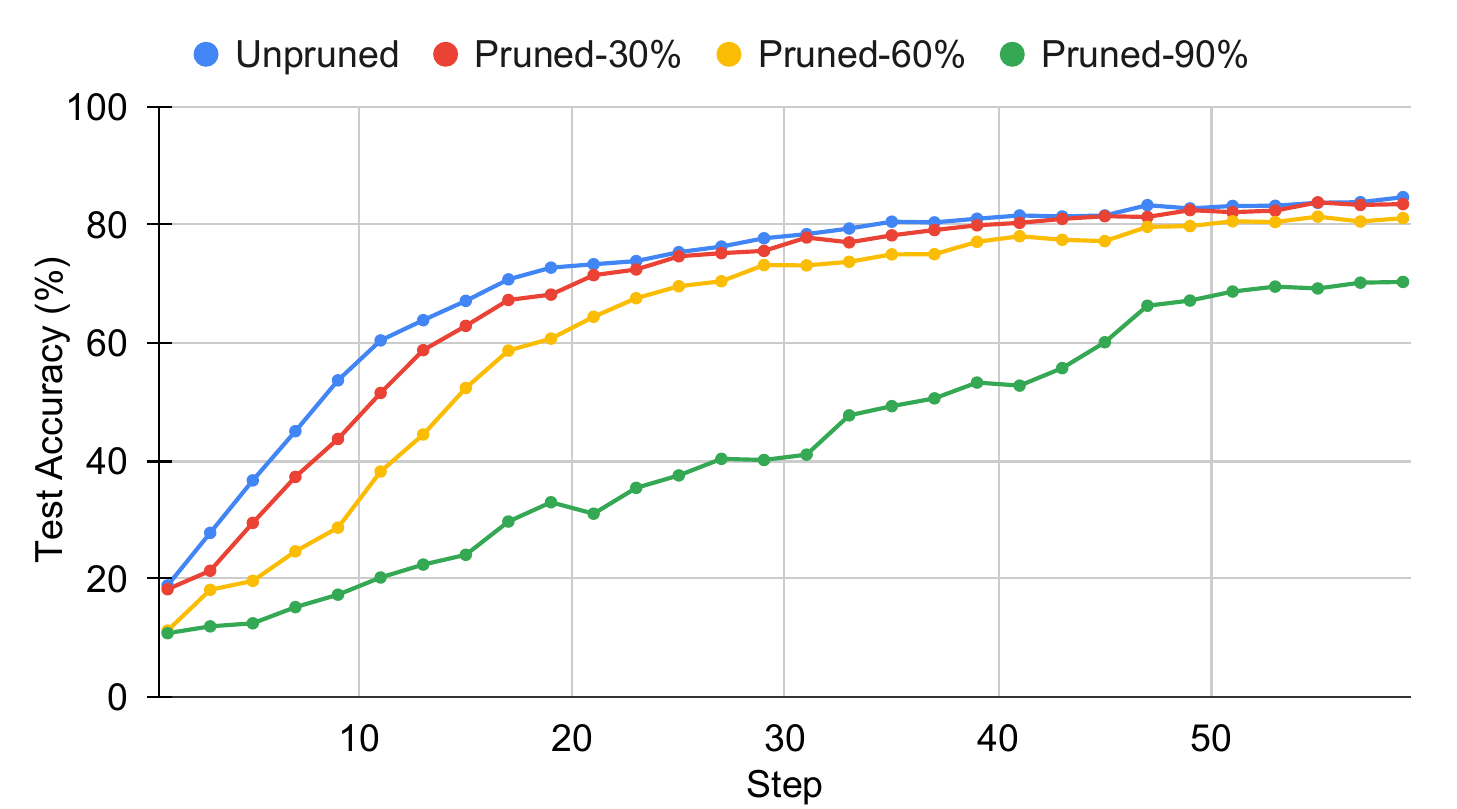}\centering
    \caption{Effect of pruning on test accuracy of VGG-16.}\label{fig:test-accuracy}
\end{figure}

For comparison, our proposed REFT employs three pruning levels: 30\%, 60\%, and 90\%, based on client device FLOPs. As seen in Figure \ref{fig:gpu-utilization}, increasing pruning lowers GPU utilization from around 84\% to about 65\%. Figure \ref{fig:test-accuracy} shows accuracy variations with increased pruning. Pruning VGG-16 to 90\% accelerates training by approximately 48\%, reducing training time. In Figure \ref{fig:inference-time}, pruning to 90\% decreases inference time by roughly 30\%. This accelerates both training and inference, enhancing FL efficiency. Since REFT utilizes structured pruning, specifically based on the L1 norm (discussed in Section \ref{Variable Pruning}), to reduce model complexity, it's noteworthy that the structured pruning approach may result in a minor decline in accuracy, as illustrated in Figure \ref{fig:test-accuracy}. To retain performance, our approach avoids pruning clients capable of training unpruned models. In contrast, approaches like FL-PQSU prune all clients to the least capable hardware level, not considering individual capacities and incurring performance loss.

In Figure \ref{fig:comparision-accuracy}, the FL-PQSU method prunes the model for all clients using a level best suited for clients $c_1$ and $c_2$. While clients $c_1$ and $c_2$ achieve satisfactory model training, the hardware resources of $c_3, c_4, \text{and} c_5$ are underutilized, resulting in a decline in accuracy. This reduction in accuracy stems from the model being pruned to a higher level than what is ideally suited for their hardware, thereby leading to suboptimal overall performance. As a result, we observe that clients $c_3, c_4,$ and $c_5$ exhibit higher accuracy when our variable pruning strategy is employed compared to the static pruning methods. This observation underscores the effectiveness of our approach in tailoring the pruning level to match the specific hardware capabilities of each client. By leveraging the benefits of variable pruning, we achieve higher accuracy rates and overall improved performance

In addition to the observed improvements in resource utilization and performance, we observed RAM utilization during our experiments. We found that RAM usage ranged from 3 to 3.3 GB across different client models. This indicates that the optimal execution of our approach aligns with industry standards, as even a relatively weak client device like the Raspberry Pi 4, which features up to 8 GB of RAM, comfortably meets the minimum memory requirement, highlighting the practicality and accessibility of our proposed approach across a range of hardware configurations.

\subsection{Model Size, Computation, and Inference Time}
\begin{figure}
    \includegraphics[width=1\linewidth]{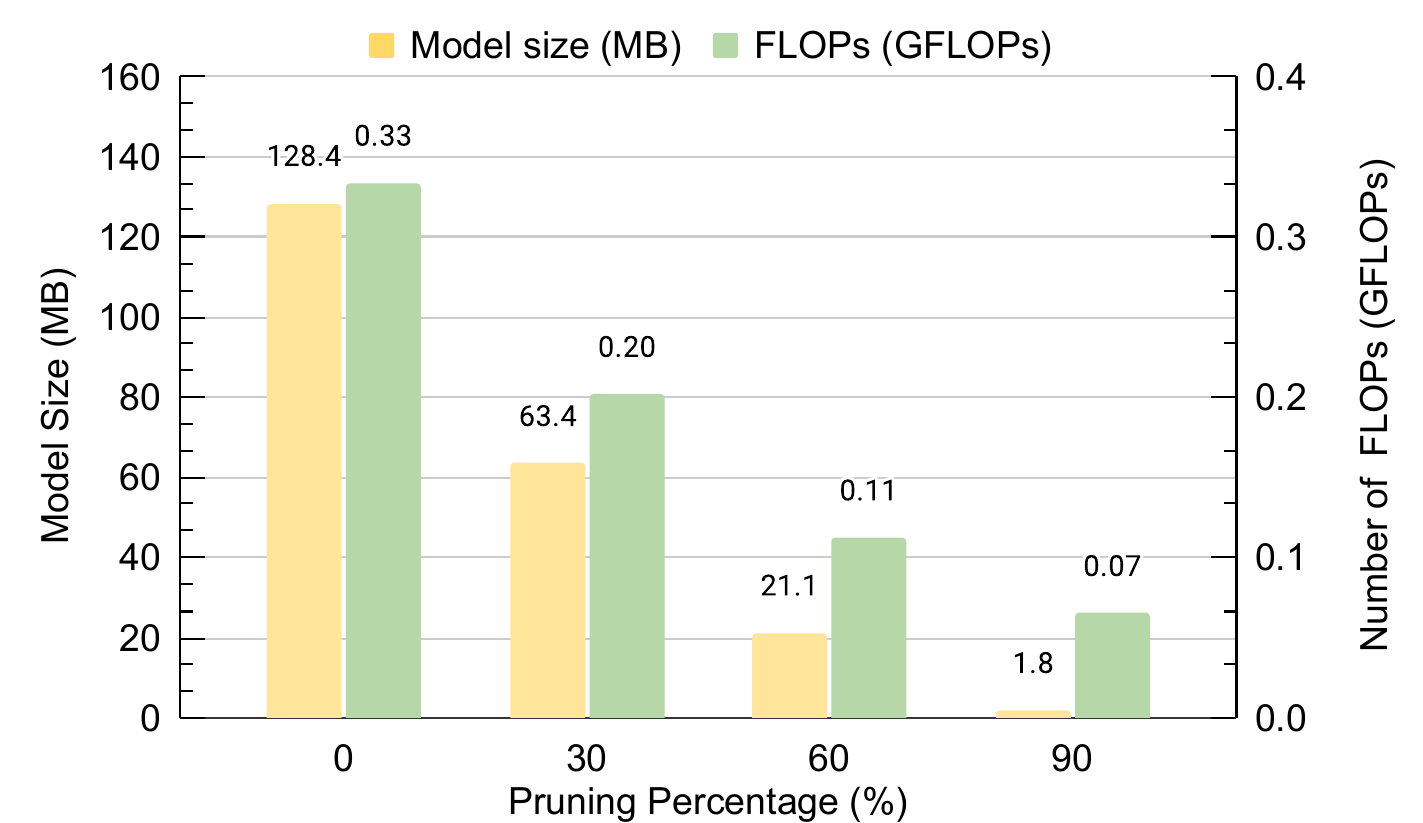}\centering
    \caption{Reduction in model size and FLOPs of VGG-16 by REFT.}\label{fig:model-size}
\end{figure}

\begin{figure}
    \includegraphics[width=1\linewidth]{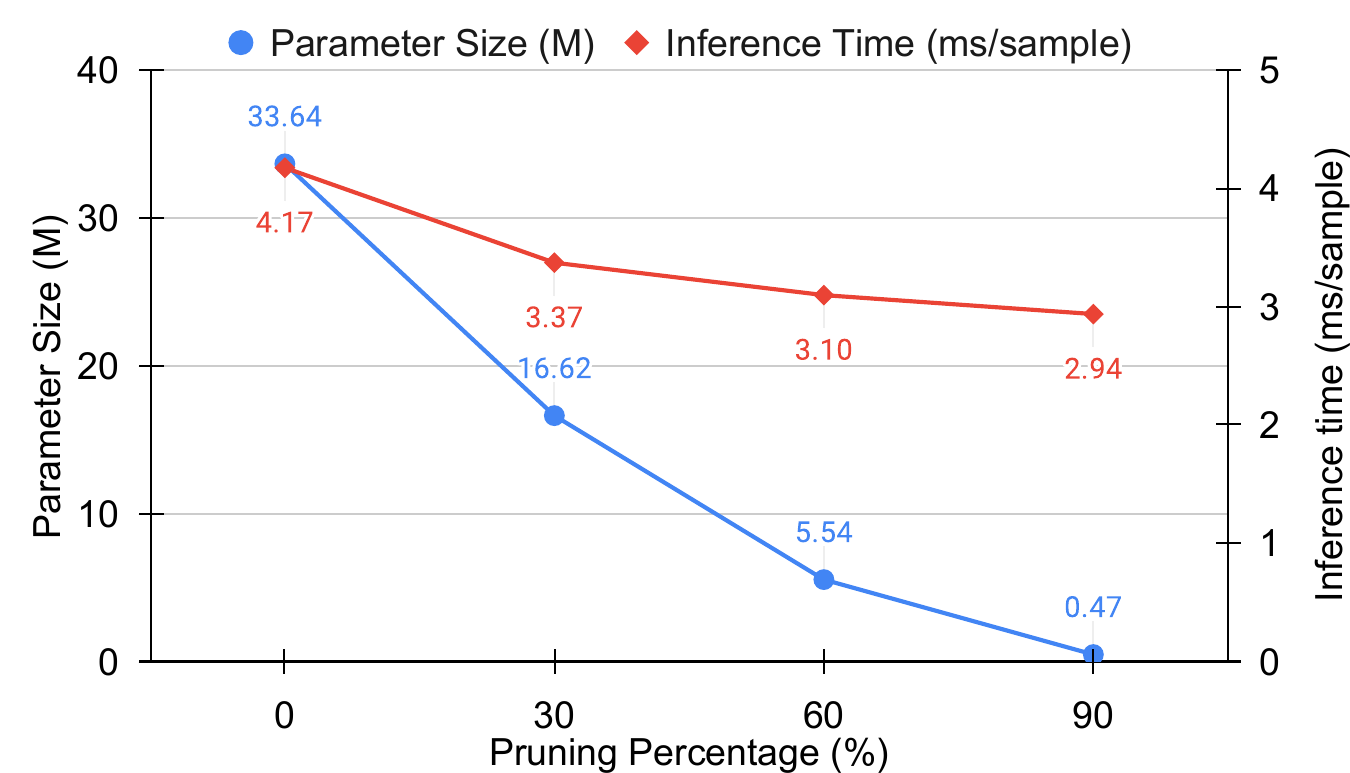}\centering
    \caption{Parameter size and inference time reduction with an increase in pruning of VGG-16 by REFT.}\label{fig:inference-time}
\end{figure}

In this set of experiments (Figures \ref{fig:model-size} and \ref{fig:inference-time}), we show that REFT achieves a substantial reduction in model size, FLOPs, and inference time as a result of the pruning performed while incurring negligible loss in final accuracy. We don't include FL-PQSU and PruneFL in the computation and inference time comparison because their pruning results in sparse matrices, which require special hardware and software for computing actual computation and inference time, as discussed in Section \ref{Model Compression}. Furthermore, PruneFL prunes the model iteratively while training until the training converges, making it unsuitable for direct comparison with specific pruning levels.
To assess the impact of REFT on model size and computation, we measured the number of model parameters and FLOPs of VGG-16 across different pruning levels. Figure \ref{fig:model-size} illustrates the significant reduction achieved by REFT, compressing the size of the model from 128.4 MB to 1.8 MB, resulting in a 98.5\% reduction. Similarly, the FLOPs are reduced from 0.33 GFLOPs to 0.07 GFLOPs.

As pruning also improves the model speedups by reducing the inference time, we visualize the improvements in inference time in Figure \ref{fig:inference-time} along with the parameter size reduction. Our pruning reduces the parameter size of VGG-16 from 33.6 M to 0.47 M parameters, achieving about 98.3\% reduction. Similarly, the inference time is reduced from 4.17 ms/sample to 2.94 ms/sample for 1000 samples.

\begin{table*}[]
\centering
\caption{Comparison of accuracy (central) and communication efficiency on RestNet-8 and VGG-16 models with 20 clients ($C = 20$). The table presents the downstream and upstream communication costs per client for each round, excluding FedKD and REFT. The total column represents the total bandwidth cost for a client to complete the federated learning training.}
\label{Table:Communication-efficiency}
\begin{tabular}{|l|c|c|c|ccc|}
\hline
                                 &                                   &                                               &                                          & \multicolumn{3}{c|}{\textbf{Bandwidth}}                                                                                                  \\ \cline{5-7} 
\multirow{-2}{*}{\textbf{Model}} & \multirow{-2}{*}{\textbf{Method}} & \multirow{-2}{*}{\textbf{Pruning Ratio (\%)}} & \multirow{-2}{*}{\textbf{Accuracy (\%)}} & \multicolumn{1}{c|}{\textbf{Downstream}}              & \multicolumn{1}{c|}{\textbf{Upstream}}                & \textbf{Total}           \\ \hline
                                 & FedAvg                            & -                                             & 68.3                                     & \multicolumn{1}{c|}{37.63 MB}                         & \multicolumn{1}{l|}{37.63 MB} & 73.5 GB                  \\ \cline{2-7} 
                                 &                                   & 30\%                                          & 73.7                                     & \multicolumn{1}{c|}{37.63 MB}                         & \multicolumn{1}{c|}{\textbf{4.7 MB}}                           & 164 GB                   \\ \cline{3-7} 
                                 &                                   & 60\%                                          & 71.9                                     & \multicolumn{1}{c|}{37.63 MB}                         & \multicolumn{1}{c|}{\textbf{4.7 MB}}                           & 164 GB                   \\ \cline{3-7} 
                                 & \multirow{-3}{*}{FL-PQSU}         & 90\%                                          & 70.3                                     & \multicolumn{1}{c|}{37.63 MB}                         & \multicolumn{1}{c|}{\textbf{4.7 MB}}                           & 164 GB                   \\ \cline{2-7} 
                                 & PruneFL                           & Adaptive ($\sim$60\%)                         & 76.2                                     & \multicolumn{1}{c|}{$\sim$20.8 MB}                          & \multicolumn{1}{c|}{$\sim$20.8 MB}                          & $\sim$36.6 GB                  \\ \cline{2-7} 
                                 & FedKD                             & -                                             & 81.5                                    & \multicolumn{1}{c|}{37.63 MB} & \multicolumn{1}{c|}{4.92 MB}                 & 42.55 MB                 \\ \cline{2-7} 
\multirow{-7}{*}{ResNet-8}       & REFT                          & Variable (30-90\%)                            & 80.8 - \bf{81.7}                            & \multicolumn{1}{c|}{\textbf{3.65 - 25.74 MB}}                  & \multicolumn{1}{c|}{7.81 MB}                          & \textbf{11.46 - 33.5 MB} \\ \hline
                                 & FedAvg                            & -                                             & 59.3                                     & \multicolumn{1}{c|}{256.6 MB}                         & \multicolumn{1}{c|}{256.6 MB}                         & 200 GB                   \\ \cline{2-7} 
                                 &                                   & 30\%                                          & 74.5                                     & \multicolumn{1}{c|}{256.6 MB} & \multicolumn{1}{c|}{32 MB}    & 140.1 GB                 \\ \cline{3-7} 
                                 &                                   & 60\%                                          & 73.1                                     & \multicolumn{1}{c|}{256.6 MB} & \multicolumn{1}{c|}{32 MB}    & 140.1 GB                 \\ \cline{3-7} 
                                 & \multirow{-3}{*}{FL-PQSU}         & 90\%                                          & 71.3                                     & \multicolumn{1}{c|}{256.6 MB} & \multicolumn{1}{c|}{32 MB}    & 140.1 GB                 \\ \cline{2-7} 
                                 & PruneFL                           & Adaptive ($\sim$60\%)                         & 78.9                                     & \multicolumn{1}{c|}{$\sim$102 MB}                     & \multicolumn{1}{c|}{$\sim$102 MB}                     & $\sim$93.6 GB            \\ \cline{2-7} 
                                 & FedKD                             & -                                             & 79.3                                     & \multicolumn{1}{c|}{256.6 MB} & \multicolumn{1}{c|}{\textbf{4.92 MB}}                          & 261.5 MB                 \\ \cline{2-7} 
\multirow{-7}{*}{VGG-16}         & REFT                          & Variable (30-90\%)                            & 77.6 - \textbf{80.8}                              & \multicolumn{1}{c|}{\textbf{3.6 - 126.8 MB}}                   & \multicolumn{1}{c|}{7.81 MB}                           & \textbf{11.4 - 134.6 MB}          \\ \hline
\end{tabular}
\end{table*}

\subsection{Communication Efficiency}
We evaluated REFT's communication efficiency against baselines using ResNet-8 and VGG-16 models. Table \ref{Table:Communication-efficiency} summarizes the outcomes in terms of total communication bandwidth and test accuracy. Notably, REFT achieves the lowest total communication bandwidth with comparable or superior accuracy compared to baselines. Due to variable pruning in REFT, accuracy and bandwidth are presented as a range to reflect pruned clients and ratios. The range accommodates various scenarios, from highly constrained clients (90\% pruning) to less constrained ones (30\% pruning). This captures REFT's trade-off between accuracy and bandwidth efficiency.


 For FedKD and REFT, the knowledge distillation process required minimal communication cost, as only 200 distillation steps were performed. In contrast, FedAvg, FL-PQSU, and PruneFL required significantly more communication rounds (900 to 1000 for ResNet-8 and about 500 for VGG-16) to complete the federated learning training. Moreover, given the dataset that we are using, the cost associated with transferring logits is considerably less than that associated with transferring model weights. Therefore, the total bandwidth of REFT and FedKD is far less than the other baselines. Given that PruneFL performs iterative pruning and reconfiguration at fixed intervals, we simplify the analysis by considering its final pruning ratio. Consequently, we provide approximate values for the bandwidth to maintain clarity and simplicity in the comparison.

 FL-PQSU stands out in terms of upstream cost reduction among weight-sharing methods like FedAvg and PruneFL. This advantage can be attributed to its utilization of INT8 quantization during client-server communication. As elaborated in Section \ref{Model Compression}, FL-PQSU maintains consistent total bandwidth usage across various pruning levels, primarily because it avoids the need for post-pruning model weight reconfiguration. Consequently, this approach leads to the emergence of sparse weight tensors and a model architecture that remains unaltered. As a result, the actual model size and computational requirements experienced no meaningful change.

Table \ref{Table:Communication-efficiency} highlights our method's superior downstream communication efficiency for both ResNet-8 and VGG-16 models, outperforming other pruning-based techniques like FL-PQSU and PruneFL. These methods anticipate sparse matrix support in future hardware and software developments, but such resources are not yet available. As a result, they encounter practical constraints, particularly PruneFL, which relies heavily on sparse matrices. In contrast, REFT adopts structured pruning, yielding hardware-friendly weight matrices that undergo size reduction through shape reconfiguration. This strategy significantly reduces parameter size and enhances latency. Furthermore, REFT and FedKD exhibit lower bandwidth needs than other baselines, leveraging one-shot distillation. Remarkably, FedKD employs quantization before transmitting logits to the server, further curbing upstream communication costs.

\section{Conclusion and Future Work}
In this study, we introduced REFT, a framework that improves resource utilization in federated learning. Through variable pruning and knowledge distillation, we enhanced resource utilization, training time, and bandwidth consumption. Our experiments showcased REFT's efficiency compared to baselines while sustaining performance. The adaptation of variable pruning enabled diverse clients to contribute effectively in resource-constrained scenarios, thereby enhancing resource utilization. Furthermore, employing one-shot knowledge distillation on public data minimizes repetitive communication, ensures privacy, and boosts efficiency. Our future work entails exploring the synergistic effect of quantization with REFT to further reduce upstream bandwidth communication, thereby enhancing overall communication efficiency.


\bibliographystyle{ACM-Reference-Format}
\bibliography{references}


\begin{thebibliography}{41}


\ifx \showCODEN    \undefined \def \showCODEN     #1{\unskip}     \fi
\ifx \showDOI      \undefined \def \showDOI       #1{#1}\fi
\ifx \showISBNx    \undefined \def \showISBNx     #1{\unskip}     \fi
\ifx \showISBNxiii \undefined \def \showISBNxiii  #1{\unskip}     \fi
\ifx \showISSN     \undefined \def \showISSN      #1{\unskip}     \fi
\ifx \showLCCN     \undefined \def \showLCCN      #1{\unskip}     \fi
\ifx \shownote     \undefined \def \shownote      #1{#1}          \fi
\ifx \showarticletitle \undefined \def \showarticletitle #1{#1}   \fi
\ifx \showURL      \undefined \def \showURL       {\relax}        \fi
\providecommand\bibfield[2]{#2}
\providecommand\bibinfo[2]{#2}
\providecommand\natexlab[1]{#1}
\providecommand\showeprint[2][]{arXiv:#2}

\bibitem[Cleland et~al\mbox{.}(2022)]%
        {FedComm}
\bibfield{author}{\bibinfo{person}{Gary Cleland}, \bibinfo{person}{Di Wu}, \bibinfo{person}{Rehmat Ullah}, {and} \bibinfo{person}{Blesson Varghese}.} \bibinfo{year}{2022}\natexlab{}.
\newblock \bibinfo{title}{FedComm: Understanding Communication Protocols for Edge-based Federated Learning}.
\newblock
\newblock
\showeprint[arxiv]{2208.08764}~[cs.DC]


\bibitem[Fallah et~al\mbox{.}(2020)]%
        {PerFedAvg}
\bibfield{author}{\bibinfo{person}{Alireza Fallah}, \bibinfo{person}{Aryan Mokhtari}, {and} \bibinfo{person}{Asuman~E. Ozdaglar}.} \bibinfo{year}{2020}\natexlab{}.
\newblock \showarticletitle{Personalized Federated Learning: {A} Meta-Learning Approach}.
\newblock \bibinfo{journal}{\emph{CoRR}}  \bibinfo{volume}{abs/2002.07948} (\bibinfo{year}{2020}).
\newblock
\showeprint[arXiv]{2002.07948}
\urldef\tempurl%
\url{https://arxiv.org/abs/2002.07948}
\showURL{%
\tempurl}


\bibitem[Gong et~al\mbox{.}(2021)]%
        {gongICCV}
\bibfield{author}{\bibinfo{person}{Xuan Gong}, \bibinfo{person}{Abhishek Sharma}, \bibinfo{person}{Srikrishna Karanam}, \bibinfo{person}{Ziyan Wu}, \bibinfo{person}{Terrence Chen}, \bibinfo{person}{David Doermann}, {and} \bibinfo{person}{Arun Innanje}.} \bibinfo{year}{2021}\natexlab{}.
\newblock \showarticletitle{Ensemble Attention Distillation for Privacy-Preserving Federated Learning}. In \bibinfo{booktitle}{\emph{2021 IEEE/CVF International Conference on Computer Vision (ICCV)}}. \bibinfo{pages}{15056--15066}.
\newblock
\urldef\tempurl%
\url{https://doi.org/10.1109/ICCV48922.2021.01480}
\showDOI{\tempurl}


\bibitem[Gong et~al\mbox{.}(2022a)]%
        {FedKD}
\bibfield{author}{\bibinfo{person}{Xuan Gong}, \bibinfo{person}{Abhishek Sharma}, \bibinfo{person}{Srikrishna Karanam}, \bibinfo{person}{Ziyan Wu}, \bibinfo{person}{Terrence Chen}, \bibinfo{person}{David~S. Doermann}, {and} \bibinfo{person}{Arun Innanje}.} \bibinfo{year}{2022}\natexlab{a}.
\newblock \showarticletitle{Preserving Privacy in Federated Learning with Ensemble Cross-Domain Knowledge Distillation}. In \bibinfo{booktitle}{\emph{Thirty-Sixth {AAAI} Conference on Artificial Intelligence, {AAAI} 2022, Thirty-Fourth Conference on Innovative Applications of Artificial Intelligence, {IAAI} 2022, The Twelveth Symposium on Educational Advances in Artificial Intelligence, {EAAI} 2022 Virtual Event, February 22 - March 1, 2022}}. \bibinfo{publisher}{{AAAI} Press}, \bibinfo{pages}{11891--11899}.
\newblock
\urldef\tempurl%
\url{https://ojs.aaai.org/index.php/AAAI/article/view/21446}
\showURL{%
\tempurl}


\bibitem[Gong et~al\mbox{.}(2022b)]%
        {gong2022FedAD}
\bibfield{author}{\bibinfo{person}{Xuan Gong}, \bibinfo{person}{Liangchen Song}, \bibinfo{person}{Rishi Vedula}, \bibinfo{person}{Abhishek Sharma}, \bibinfo{person}{Meng Zheng}, \bibinfo{person}{Benjamin Planche}, \bibinfo{person}{Arun Innanje}, \bibinfo{person}{Terrence Chen}, \bibinfo{person}{Junsong Yuan}, \bibinfo{person}{David Doermann}, {and} \bibinfo{person}{Ziyan Wu}.} \bibinfo{year}{2022}\natexlab{b}.
\newblock \bibinfo{title}{Federated Learning with Privacy-Preserving Ensemble Attention Distillation}.
\newblock
\newblock
\showeprint[arxiv]{2210.08464}~[cs.LG]


\bibitem[Goodfellow et~al\mbox{.}(2014)]%
        {GAN}
\bibfield{author}{\bibinfo{person}{Ian~J. Goodfellow}, \bibinfo{person}{Jean Pouget{-}Abadie}, \bibinfo{person}{Mehdi Mirza}, \bibinfo{person}{Bing Xu}, \bibinfo{person}{David Warde{-}Farley}, \bibinfo{person}{Sherjil Ozair}, \bibinfo{person}{Aaron~C. Courville}, {and} \bibinfo{person}{Yoshua Bengio}.} \bibinfo{year}{2014}\natexlab{}.
\newblock \showarticletitle{Generative Adversarial Networks}.
\newblock \bibinfo{journal}{\emph{CoRR}}  \bibinfo{volume}{abs/1406.2661} (\bibinfo{year}{2014}).
\newblock
\showeprint[arXiv]{1406.2661}
\urldef\tempurl%
\url{http://arxiv.org/abs/1406.2661}
\showURL{%
\tempurl}


\bibitem[Guha et~al\mbox{.}(2019)]%
        {Guha2019}
\bibfield{author}{\bibinfo{person}{Neel Guha}, \bibinfo{person}{Ameet Talwalkar}, {and} \bibinfo{person}{Virginia Smith}.} \bibinfo{year}{2019}\natexlab{}.
\newblock \showarticletitle{One-Shot Federated Learning}.
\newblock \bibinfo{journal}{\emph{CoRR}}  \bibinfo{volume}{abs/1902.11175} (\bibinfo{year}{2019}).
\newblock
\showeprint[arXiv]{1902.11175}
\urldef\tempurl%
\url{http://arxiv.org/abs/1902.11175}
\showURL{%
\tempurl}


\bibitem[Han et~al\mbox{.}(2016)]%
        {DeepCompression}
\bibfield{author}{\bibinfo{person}{Song Han}, \bibinfo{person}{Huizi Mao}, {and} \bibinfo{person}{William~J. Dally}.} \bibinfo{year}{2016}\natexlab{}.
\newblock \showarticletitle{Deep Compression: Compressing Deep Neural Network with Pruning, Trained Quantization and Huffman Coding}. In \bibinfo{booktitle}{\emph{4th International Conference on Learning Representations, {ICLR} 2016, San Juan, Puerto Rico, May 2-4, 2016, Conference Track Proceedings}}, \bibfield{editor}{\bibinfo{person}{Yoshua Bengio} {and} \bibinfo{person}{Yann LeCun}} (Eds.).
\newblock
\urldef\tempurl%
\url{http://arxiv.org/abs/1510.00149}
\showURL{%
\tempurl}


\bibitem[Han et~al\mbox{.}(2015)]%
        {Han2015}
\bibfield{author}{\bibinfo{person}{Song Han}, \bibinfo{person}{Jeff Pool}, \bibinfo{person}{John Tran}, {and} \bibinfo{person}{William Dally}.} \bibinfo{year}{2015}\natexlab{}.
\newblock \showarticletitle{Learning both Weights and Connections for Efficient Neural Network}. In \bibinfo{booktitle}{\emph{Advances in Neural Information Processing Systems}}, \bibfield{editor}{\bibinfo{person}{C.~Cortes}, \bibinfo{person}{N.~Lawrence}, \bibinfo{person}{D.~Lee}, \bibinfo{person}{M.~Sugiyama}, {and} \bibinfo{person}{R.~Garnett}} (Eds.), Vol.~\bibinfo{volume}{28}. \bibinfo{publisher}{Curran Associates, Inc.}
\newblock
\urldef\tempurl%
\url{https://proceedings.neurips.cc/paper_files/paper/2015/file/ae0eb3eed39d2bcef4622b2499a05fe6-Paper.pdf}
\showURL{%
\tempurl}


\bibitem[Hassibi et~al\mbox{.}(1993)]%
        {Hassibi1993}
\bibfield{author}{\bibinfo{person}{Babak Hassibi}, \bibinfo{person}{David Stork}, {and} \bibinfo{person}{Gregory Wolff}.} \bibinfo{year}{1993}\natexlab{}.
\newblock \showarticletitle{Optimal Brain Surgeon: Extensions and performance comparisons}. In \bibinfo{booktitle}{\emph{Advances in Neural Information Processing Systems}}, \bibfield{editor}{\bibinfo{person}{J.~Cowan}, \bibinfo{person}{G.~Tesauro}, {and} \bibinfo{person}{J.~Alspector}} (Eds.), Vol.~\bibinfo{volume}{6}. \bibinfo{publisher}{Morgan-Kaufmann}.
\newblock
\urldef\tempurl%
\url{https://proceedings.neurips.cc/paper_files/paper/1993/file/b056eb1587586b71e2da9acfe4fbd19e-Paper.pdf}
\showURL{%
\tempurl}


\bibitem[Hinton et~al\mbox{.}(2015)]%
        {hinton2015distilling}
\bibfield{author}{\bibinfo{person}{Geoffrey Hinton}, \bibinfo{person}{Oriol Vinyals}, {and} \bibinfo{person}{Jeff Dean}.} \bibinfo{year}{2015}\natexlab{}.
\newblock \bibinfo{title}{Distilling the Knowledge in a Neural Network}.
\newblock
\newblock
\showeprint[arxiv]{1503.02531}~[stat.ML]


\bibitem[Hsu et~al\mbox{.}(2019)]%
        {non-iid-data}
\bibfield{author}{\bibinfo{person}{Tzu{-}Ming~Harry Hsu}, \bibinfo{person}{Hang Qi}, {and} \bibinfo{person}{Matthew Brown}.} \bibinfo{year}{2019}\natexlab{}.
\newblock \showarticletitle{Measuring the Effects of Non-Identical Data Distribution for Federated Visual Classification}.
\newblock \bibinfo{journal}{\emph{CoRR}}  \bibinfo{volume}{abs/1909.06335} (\bibinfo{year}{2019}).
\newblock
\showeprint[arXiv]{1909.06335}
\urldef\tempurl%
\url{http://arxiv.org/abs/1909.06335}
\showURL{%
\tempurl}


\bibitem[Hsu et~al\mbox{.}(2020)]%
        {FedVC}
\bibfield{author}{\bibinfo{person}{Tzu{-}Ming~Harry Hsu}, \bibinfo{person}{Hang Qi}, {and} \bibinfo{person}{Matthew Brown}.} \bibinfo{year}{2020}\natexlab{}.
\newblock \showarticletitle{Federated Visual Classification with Real-World Data Distribution}.
\newblock \bibinfo{journal}{\emph{CoRR}}  \bibinfo{volume}{abs/2003.08082} (\bibinfo{year}{2020}).
\newblock
\showeprint[arXiv]{2003.08082}
\urldef\tempurl%
\url{https://arxiv.org/abs/2003.08082}
\showURL{%
\tempurl}


\bibitem[Jiang et~al\mbox{.}(2019)]%
        {PruneFL}
\bibfield{author}{\bibinfo{person}{Yuang Jiang}, \bibinfo{person}{Shiqiang Wang}, \bibinfo{person}{Bong~Jun Ko}, \bibinfo{person}{Wei{-}Han Lee}, {and} \bibinfo{person}{Leandros Tassiulas}.} \bibinfo{year}{2019}\natexlab{}.
\newblock \showarticletitle{Model Pruning Enables Efficient Federated Learning on Edge Devices}.
\newblock \bibinfo{journal}{\emph{CoRR}}  \bibinfo{volume}{abs/1909.12326} (\bibinfo{year}{2019}).
\newblock
\showeprint[arXiv]{1909.12326}
\urldef\tempurl%
\url{http://arxiv.org/abs/1909.12326}
\showURL{%
\tempurl}


\bibitem[Kone{\v{c}}n{\'y} et~al\mbox{.}(2016)]%
        {sketched-updateFL}
\bibfield{author}{\bibinfo{person}{Jakub Kone{\v{c}}n{\'y}}, \bibinfo{person}{H.~Brendan McMahan}, \bibinfo{person}{Felix~X. Yu}, \bibinfo{person}{Peter Richt{\'{a}}rik}, \bibinfo{person}{Ananda~Theertha Suresh}, {and} \bibinfo{person}{Dave Bacon}.} \bibinfo{year}{2016}\natexlab{}.
\newblock \showarticletitle{Federated Learning: Strategies for Improving Communication Efficiency}.
\newblock \bibinfo{journal}{\emph{CoRR}}  \bibinfo{volume}{abs/1610.05492} (\bibinfo{year}{2016}).
\newblock
\showeprint[arXiv]{1610.05492}
\urldef\tempurl%
\url{http://arxiv.org/abs/1610.05492}
\showURL{%
\tempurl}


\bibitem[Krizhevsky(2009)]%
        {krizhevsky2009learning}
\bibfield{author}{\bibinfo{person}{Alex Krizhevsky}.} \bibinfo{year}{2009}\natexlab{}.
\newblock \showarticletitle{Learning Multiple Layers of Features from Tiny Images}.
\newblock  (\bibinfo{year}{2009}), \bibinfo{pages}{32--33}.
\newblock
\urldef\tempurl%
\url{https://www.cs.toronto.edu/~kriz/learning-features-2009-TR.pdf}
\showURL{%
\tempurl}


\bibitem[Lecun et~al\mbox{.}(1990)]%
        {LeChun1990}
\bibfield{author}{\bibinfo{person}{Yann Lecun}, \bibinfo{person}{{J. S.} Denker}, \bibinfo{person}{{Sara A.} Solla}, \bibinfo{person}{{R. E.} Howard}, {and} \bibinfo{person}{L.D. Jackel}.} \bibinfo{year}{1990}\natexlab{}.
\newblock \showarticletitle{Optimal brain damage}. In \bibinfo{booktitle}{\emph{Advances in Neural Information Processing Systems (NIPS 1989), Denver, CO}}, \bibfield{editor}{\bibinfo{person}{David Touretzky}} (Ed.), Vol.~\bibinfo{volume}{2}. \bibinfo{publisher}{Morgan Kaufmann}.
\newblock


\bibitem[Li et~al\mbox{.}(2017)]%
        {li2017pruning}
\bibfield{author}{\bibinfo{person}{Hao Li}, \bibinfo{person}{Asim Kadav}, \bibinfo{person}{Igor Durdanovic}, \bibinfo{person}{Hanan Samet}, {and} \bibinfo{person}{Hans~Peter Graf}.} \bibinfo{year}{2017}\natexlab{}.
\newblock \showarticletitle{Pruning Filters for Efficient ConvNets}. In \bibinfo{booktitle}{\emph{International Conference on Learning Representations}}.
\newblock
\urldef\tempurl%
\url{https://openreview.net/forum?id=rJqFGTslg}
\showURL{%
\tempurl}


\bibitem[Li et~al\mbox{.}(2019a)]%
        {FLChallenges}
\bibfield{author}{\bibinfo{person}{Tian Li}, \bibinfo{person}{Anit~Kumar Sahu}, \bibinfo{person}{Ameet Talwalkar}, {and} \bibinfo{person}{Virginia Smith}.} \bibinfo{year}{2019}\natexlab{a}.
\newblock \showarticletitle{Federated Learning: Challenges, Methods, and Future Directions}.
\newblock \bibinfo{journal}{\emph{CoRR}}  \bibinfo{volume}{abs/1908.07873} (\bibinfo{year}{2019}).
\newblock
\showeprint[arXiv]{1908.07873}
\urldef\tempurl%
\url{http://arxiv.org/abs/1908.07873}
\showURL{%
\tempurl}


\bibitem[Li et~al\mbox{.}(2020)]%
        {FairAllocFL}
\bibfield{author}{\bibinfo{person}{Tian Li}, \bibinfo{person}{Maziar Sanjabi}, \bibinfo{person}{Ahmad Beirami}, {and} \bibinfo{person}{Virginia Smith}.} \bibinfo{year}{2020}\natexlab{}.
\newblock \bibinfo{title}{Fair Resource Allocation in Federated Learning}.
\newblock
\newblock
\showeprint[arxiv]{1905.10497}~[cs.LG]


\bibitem[Li et~al\mbox{.}(2019b)]%
        {q-fedAvg}
\bibfield{author}{\bibinfo{person}{Tian Li}, \bibinfo{person}{Maziar Sanjabi}, {and} \bibinfo{person}{Virginia Smith}.} \bibinfo{year}{2019}\natexlab{b}.
\newblock \showarticletitle{Fair Resource Allocation in Federated Learning}.
\newblock \bibinfo{journal}{\emph{CoRR}}  \bibinfo{volume}{abs/1905.10497} (\bibinfo{year}{2019}).
\newblock
\showeprint[arXiv]{1905.10497}
\urldef\tempurl%
\url{http://arxiv.org/abs/1905.10497}
\showURL{%
\tempurl}


\bibitem[Lin et~al\mbox{.}(2020)]%
        {FedDF}
\bibfield{author}{\bibinfo{person}{Tao Lin}, \bibinfo{person}{Lingjing Kong}, \bibinfo{person}{Sebastian~U. Stich}, {and} \bibinfo{person}{Martin Jaggi}.} \bibinfo{year}{2020}\natexlab{}.
\newblock \showarticletitle{Ensemble Distillation for Robust Model Fusion in Federated Learning}. In \bibinfo{booktitle}{\emph{Advances in Neural Information Processing Systems 33: Annual Conference on Neural Information Processing Systems 2020, NeurIPS 2020, December 6-12, 2020, virtual}}, \bibfield{editor}{\bibinfo{person}{Hugo Larochelle}, \bibinfo{person}{Marc'Aurelio Ranzato}, \bibinfo{person}{Raia Hadsell}, \bibinfo{person}{Maria{-}Florina Balcan}, {and} \bibinfo{person}{Hsuan{-}Tien Lin}} (Eds.).
\newblock
\urldef\tempurl%
\url{https://proceedings.neurips.cc/paper/2020/hash/18df51b97ccd68128e994804f3eccc87-Abstract.html}
\showURL{%
\tempurl}


\bibitem[McMahan et~al\mbox{.}(2023)]%
        {McMahan}
\bibfield{author}{\bibinfo{person}{H.~Brendan McMahan}, \bibinfo{person}{Eider Moore}, \bibinfo{person}{Daniel Ramage}, \bibinfo{person}{Seth Hampson}, {and} \bibinfo{person}{Blaise~Agüera y Arcas}.} \bibinfo{year}{2023}\natexlab{}.
\newblock \bibinfo{title}{Communication-Efficient Learning of Deep Networks from Decentralized Data}.
\newblock
\newblock
\showeprint[arxiv]{1602.05629}~[cs.LG]


\bibitem[Micaelli and Storkey(2019)]%
        {Zero-shotKTviaBM}
\bibfield{author}{\bibinfo{person}{Paul Micaelli} {and} \bibinfo{person}{Amos~J. Storkey}.} \bibinfo{year}{2019}\natexlab{}.
\newblock \showarticletitle{Zero-shot Knowledge Transfer via Adversarial Belief Matching}. In \bibinfo{booktitle}{\emph{Advances in Neural Information Processing Systems 32: Annual Conference on Neural Information Processing Systems 2019, NeurIPS 2019, December 8-14, 2019, Vancouver, BC, Canada}}, \bibfield{editor}{\bibinfo{person}{Hanna~M. Wallach}, \bibinfo{person}{Hugo Larochelle}, \bibinfo{person}{Alina Beygelzimer}, \bibinfo{person}{Florence d'Alch{\'{e}}{-}Buc}, \bibinfo{person}{Emily~B. Fox}, {and} \bibinfo{person}{Roman Garnett}} (Eds.). \bibinfo{pages}{9547--9557}.
\newblock
\urldef\tempurl%
\url{https://proceedings.neurips.cc/paper/2019/hash/fe663a72b27bdc613873fbbb512f6f67-Abstract.html}
\showURL{%
\tempurl}


\bibitem[{Microsoft NNI Contributors}(2023)]%
        {nniOverviewModel}
\bibfield{author}{\bibinfo{person}{{Microsoft NNI Contributors}}.} \bibinfo{year}{2023}\natexlab{}.
\newblock \bibinfo{title}{{O}verview of {N}{N}{I} {M}odel {P}runing; {N}eural {N}etwork {I}ntelligence --- nni.readthedocs.io}.
\newblock
\newblock
\urldef\tempurl%
\url{https://nni.readthedocs.io/en/stable/compression/pruning.html#dependency-aware-mode-for-output-channel-pruning}
\showURL{%
\tempurl}
\newblock
\shownote{[Accessed 14-08-2023]}.


\bibitem[Mohri et~al\mbox{.}(2019)]%
        {AgnosticFL}
\bibfield{author}{\bibinfo{person}{Mehryar Mohri}, \bibinfo{person}{Gary Sivek}, {and} \bibinfo{person}{Ananda~Theertha Suresh}.} \bibinfo{year}{2019}\natexlab{}.
\newblock \bibinfo{title}{Agnostic Federated Learning}.
\newblock
\newblock
\showeprint[arxiv]{1902.00146}~[cs.LG]


\bibitem[Nayak et~al\mbox{.}(2019)]%
        {Zero-ShotKD}
\bibfield{author}{\bibinfo{person}{Gaurav~Kumar Nayak}, \bibinfo{person}{Konda~Reddy Mopuri}, \bibinfo{person}{Vaisakh Shaj}, \bibinfo{person}{R.~Venkatesh Babu}, {and} \bibinfo{person}{Anirban Chakraborty}.} \bibinfo{year}{2019}\natexlab{}.
\newblock \showarticletitle{Zero-Shot Knowledge Distillation in Deep Networks}.
\newblock \bibinfo{journal}{\emph{CoRR}}  \bibinfo{volume}{abs/1905.08114} (\bibinfo{year}{2019}).
\newblock
\showeprint[arXiv]{1905.08114}
\urldef\tempurl%
\url{http://arxiv.org/abs/1905.08114}
\showURL{%
\tempurl}


\bibitem[Reisizadeh et~al\mbox{.}(2020)]%
        {FedPAQ_PMLR}
\bibfield{author}{\bibinfo{person}{Amirhossein Reisizadeh}, \bibinfo{person}{Aryan Mokhtari}, \bibinfo{person}{Hamed Hassani}, \bibinfo{person}{Ali Jadbabaie}, {and} \bibinfo{person}{Ramtin Pedarsani}.} \bibinfo{year}{2020}\natexlab{}.
\newblock \showarticletitle{FedPAQ: A Communication-Efficient Federated Learning Method with Periodic Averaging and Quantization}. In \bibinfo{booktitle}{\emph{Proceedings of the Twenty Third International Conference on Artificial Intelligence and Statistics}} \emph{(\bibinfo{series}{Proceedings of Machine Learning Research}, Vol.~\bibinfo{volume}{108})}, \bibfield{editor}{\bibinfo{person}{Silvia Chiappa} {and} \bibinfo{person}{Roberto Calandra}} (Eds.). \bibinfo{publisher}{PMLR}, \bibinfo{pages}{2021--2031}.
\newblock
\urldef\tempurl%
\url{https://proceedings.mlr.press/v108/reisizadeh20a.html}
\showURL{%
\tempurl}


\bibitem[Romero et~al\mbox{.}(2015)]%
        {romero2015fitnets}
\bibfield{author}{\bibinfo{person}{Adriana Romero}, \bibinfo{person}{Nicolas Ballas}, \bibinfo{person}{Samira~Ebrahimi Kahou}, \bibinfo{person}{Antoine Chassang}, \bibinfo{person}{Carlo Gatta}, {and} \bibinfo{person}{Yoshua Bengio}.} \bibinfo{year}{2015}\natexlab{}.
\newblock \bibinfo{title}{FitNets: Hints for Thin Deep Nets}.
\newblock
\newblock
\showeprint[arxiv]{1412.6550}~[cs.LG]


\bibitem[Sattler et~al\mbox{.}(2020)]%
        {RobCommEFL}
\bibfield{author}{\bibinfo{person}{Felix Sattler}, \bibinfo{person}{Simon Wiedemann}, \bibinfo{person}{Klaus-Robert Müller}, {and} \bibinfo{person}{Wojciech Samek}.} \bibinfo{year}{2020}\natexlab{}.
\newblock \showarticletitle{Robust and Communication-Efficient Federated Learning From Non-i.i.d. Data}.
\newblock \bibinfo{journal}{\emph{IEEE Transactions on Neural Networks and Learning Systems}} \bibinfo{volume}{31}, \bibinfo{number}{9} (\bibinfo{year}{2020}), \bibinfo{pages}{3400--3413}.
\newblock
\urldef\tempurl%
\url{https://doi.org/10.1109/TNNLS.2019.2944481}
\showDOI{\tempurl}


\bibitem[Singh et~al\mbox{.}(2019)]%
        {PlayAndPrune}
\bibfield{author}{\bibinfo{person}{Pravendra Singh}, \bibinfo{person}{Vinay~Kumar Verma}, \bibinfo{person}{Piyush Rai}, {and} \bibinfo{person}{Vinay~P. Namboodiri}.} \bibinfo{year}{2019}\natexlab{}.
\newblock \showarticletitle{Play and Prune: Adaptive Filter Pruning for Deep Model Compression}.
\newblock \bibinfo{journal}{\emph{CoRR}}  \bibinfo{volume}{abs/1905.04446} (\bibinfo{year}{2019}).
\newblock
\showeprint[arXiv]{1905.04446}
\urldef\tempurl%
\url{http://arxiv.org/abs/1905.04446}
\showURL{%
\tempurl}


\bibitem[Tang et~al\mbox{.}(2018)]%
        {FLOPsRTang2018}
\bibfield{author}{\bibinfo{person}{Raphael Tang}, \bibinfo{person}{Ashutosh Adhikari}, {and} \bibinfo{person}{Jimmy Lin}.} \bibinfo{year}{2018}\natexlab{}.
\newblock \showarticletitle{FLOPs as a Direct Optimization Objective for Learning Sparse Neural Networks}.
\newblock \bibinfo{journal}{\emph{CoRR}}  \bibinfo{volume}{abs/1811.03060} (\bibinfo{year}{2018}).
\newblock
\showeprint[arXiv]{1811.03060}
\urldef\tempurl%
\url{http://arxiv.org/abs/1811.03060}
\showURL{%
\tempurl}


\bibitem[Tang et~al\mbox{.}(2017)]%
        {PowerComsumptionRTang2017}
\bibfield{author}{\bibinfo{person}{Raphael Tang}, \bibinfo{person}{Weijie Wang}, \bibinfo{person}{Zhucheng Tu}, {and} \bibinfo{person}{Jimmy Lin}.} \bibinfo{year}{2017}\natexlab{}.
\newblock \showarticletitle{An Experimental Analysis of the Power Consumption of Convolutional Neural Networks for Keyword Spotting}.
\newblock \bibinfo{journal}{\emph{CoRR}}  \bibinfo{volume}{abs/1711.00333} (\bibinfo{year}{2017}).
\newblock
\showeprint[arXiv]{1711.00333}
\urldef\tempurl%
\url{http://arxiv.org/abs/1711.00333}
\showURL{%
\tempurl}


\bibitem[Tung and Mori(2019)]%
        {Tung_2019_ICCV}
\bibfield{author}{\bibinfo{person}{Frederick Tung} {and} \bibinfo{person}{Greg Mori}.} \bibinfo{year}{2019}\natexlab{}.
\newblock \showarticletitle{Similarity-Preserving Knowledge Distillation}. In \bibinfo{booktitle}{\emph{Proceedings of the IEEE/CVF International Conference on Computer Vision (ICCV)}}.
\newblock


\bibitem[Vineeth(2021)]%
        {vineeth_2021}
\bibfield{author}{\bibinfo{person}{S Vineeth}.} \bibinfo{year}{2021}\natexlab{}.
\newblock \bibinfo{title}{Federated learning over WiFi: Should we use TCP or UDP?}
\newblock
\newblock
\urldef\tempurl%
\url{https://doi.org/10.31219/osf.io/tuz6c}
\showDOI{\tempurl}


\bibitem[Wang et~al\mbox{.}(2020)]%
        {MatchedAvg}
\bibfield{author}{\bibinfo{person}{Hongyi Wang}, \bibinfo{person}{Mikhail Yurochkin}, \bibinfo{person}{Yuekai Sun}, \bibinfo{person}{Dimitris Papailiopoulos}, {and} \bibinfo{person}{Yasaman Khazaeni}.} \bibinfo{year}{2020}\natexlab{}.
\newblock \bibinfo{title}{Federated Learning with Matched Averaging}.
\newblock
\newblock
\showeprint[arxiv]{2002.06440}~[cs.LG]


\bibitem[Wang et~al\mbox{.}(2019)]%
        {NonStruct-pruning}
\bibfield{author}{\bibinfo{person}{Yanzhi Wang}, \bibinfo{person}{Shaokai Ye}, \bibinfo{person}{Zhezhi He}, \bibinfo{person}{Xiaolong Ma}, \bibinfo{person}{Linfeng Zhang}, \bibinfo{person}{Sheng Lin}, \bibinfo{person}{Geng Yuan}, \bibinfo{person}{Sia~Huat Tan}, \bibinfo{person}{Zhengang Li}, \bibinfo{person}{Deliang Fan}, \bibinfo{person}{Xuehai Qian}, \bibinfo{person}{Xue Lin}, {and} \bibinfo{person}{Kaisheng Ma}.} \bibinfo{year}{2019}\natexlab{}.
\newblock \showarticletitle{Non-structured {DNN} Weight Pruning Considered Harmful}.
\newblock \bibinfo{journal}{\emph{CoRR}}  \bibinfo{volume}{abs/1907.02124} (\bibinfo{year}{2019}).
\newblock
\showeprint[arXiv]{1907.02124}
\urldef\tempurl%
\url{http://arxiv.org/abs/1907.02124}
\showURL{%
\tempurl}


\bibitem[Xu et~al\mbox{.}(2021)]%
        {FL-PQSU}
\bibfield{author}{\bibinfo{person}{Wenyuan Xu}, \bibinfo{person}{Weiwei Fang}, \bibinfo{person}{Yi Ding}, \bibinfo{person}{Meixia Zou}, {and} \bibinfo{person}{Naixue Xiong}.} \bibinfo{year}{2021}\natexlab{}.
\newblock \showarticletitle{Accelerating Federated Learning for IoT in Big Data Analytics With Pruning, Quantization and Selective Updating}.
\newblock \bibinfo{journal}{\emph{IEEE Access}}  \bibinfo{volume}{9} (\bibinfo{year}{2021}), \bibinfo{pages}{38457--38466}.
\newblock
\urldef\tempurl%
\url{https://doi.org/10.1109/ACCESS.2021.3063291}
\showDOI{\tempurl}


\bibitem[Yang et~al\mbox{.}(2021)]%
        {WirelessEnergyEFL}
\bibfield{author}{\bibinfo{person}{Zhaohui Yang}, \bibinfo{person}{Mingzhe Chen}, \bibinfo{person}{Walid Saad}, \bibinfo{person}{Choong~Seon Hong}, {and} \bibinfo{person}{Mohammad Shikh-Bahaei}.} \bibinfo{year}{2021}\natexlab{}.
\newblock \showarticletitle{Energy Efficient Federated Learning Over Wireless Communication Networks}.
\newblock \bibinfo{journal}{\emph{IEEE Transactions on Wireless Communications}} \bibinfo{volume}{20}, \bibinfo{number}{3} (\bibinfo{year}{2021}), \bibinfo{pages}{1935--1949}.
\newblock
\urldef\tempurl%
\url{https://doi.org/10.1109/TWC.2020.3037554}
\showDOI{\tempurl}


\bibitem[Yurochkin et~al\mbox{.}(2019)]%
        {yurochkin2019bayesian}
\bibfield{author}{\bibinfo{person}{Mikhail Yurochkin}, \bibinfo{person}{Mayank Agarwal}, \bibinfo{person}{Soumya Ghosh}, \bibinfo{person}{Kristjan Greenewald}, \bibinfo{person}{Trong~Nghia Hoang}, {and} \bibinfo{person}{Yasaman Khazaeni}.} \bibinfo{year}{2019}\natexlab{}.
\newblock \bibinfo{title}{Bayesian Nonparametric Federated Learning of Neural Networks}.
\newblock
\newblock
\showeprint[arxiv]{1905.12022}~[stat.ML]


\bibitem[Zagoruyko and Komodakis(2016)]%
        {AttentiontoAttention}
\bibfield{author}{\bibinfo{person}{Sergey Zagoruyko} {and} \bibinfo{person}{Nikos Komodakis}.} \bibinfo{year}{2016}\natexlab{}.
\newblock \showarticletitle{Paying More Attention to Attention: Improving the Performance of Convolutional Neural Networks via Attention Transfer}.
\newblock \bibinfo{journal}{\emph{CoRR}}  \bibinfo{volume}{abs/1612.03928} (\bibinfo{year}{2016}).
\newblock
\showeprint[arXiv]{1612.03928}
\urldef\tempurl%
\url{http://arxiv.org/abs/1612.03928}
\showURL{%
\tempurl}


\end{thebibliography}









\end{document}